\documentclass[runningheads]{llncs}
\usepackage[T1]{fontenc}
\usepackage{graphicx}
\usepackage{enumitem}
\usepackage{amsmath,amssymb,amsfonts}
\usepackage[table,dvipsnames]{xcolor}
\usepackage{algorithm}
\usepackage{algorithmicx}
\usepackage{algpseudocode}
\usepackage[table,xcdraw]{xcolor}
\usepackage{wrapfig}

\usepackage{tikz,xcolor,hyperref} 
\definecolor{lime}{HTML}{A6CE39}
\DeclareRobustCommand{\orcidicon}{
\begin{tikzpicture}
\draw[lime, fill=lime] (0,0) 
circle [radius=0.16] 
node[white] {{\fontfamily{qag}\selectfont \tiny ID}};    \draw[white, fill=white] (-0.0625,0.095) 
circle [radius=0.007];    \end{tikzpicture}
\hspace{-2mm}}
\foreach \x in {A, ..., Z}{
\expandafter\xdef\csname orcid\x\endcsname{\noexpand\href{https://orcid.org/\csname orcidauthor\x\endcsname}{\noexpand\orcidicon}}}

\begin{document}
\title{Hybrid unary-binary design for multiplier-less printed Machine Learning classifiers}

\titlerunning{Hybrid Unary-Binary MLPs}

\authorrunning{G. Armeniakos et al.}

%
%
\author{Giorgos Armeniakos\orcidA{} \and
Theodoros Mantzakidis\orcidB{} \and
Dimitrios Soudris\orcidC{}}
\authorrunning{G. Armeniakos et al.}
%
\institute{National and Technical University of Athens, Greece \email{\{armeniakos,mantzakidis,dsoudris\}@microlab.ntua.gr}}

\maketitle              

\begin{abstract}
Printed Electronics (PE) provide a flexible, cost-efficient alternative to silicon for implementing machine learning (ML) circuits, but their large feature sizes limit classifier complexity. Leveraging PE’s low fabrication and NRE costs, designers can tailor hardware to specific ML models, simplifying circuit design. This work explores alternative arithmetic and proposes a hybrid unary-binary architecture that removes costly encoders and enables efficient, multiplier-less execution of MLP classifiers. We also introduce architecture-aware training to further improve area and power efficiency. Evaluation on six datasets shows average reductions of 46\% in area and 39\% in power, with minimal accuracy loss—surpassing other state-of-the-art MLP designs.

\end{abstract}
\section{Introduction}

Printed electronics (PE) combine materials science, circuit design, and fabrication to enable flexible, stretchable, low-cost systems for emerging applications~\cite{shao2014fully}. In healthcare, PE powers wearable sensors and smart bandages; in logistics, smart packaging enables real-time tracking; and in diagnostics, it supports disposable, low-cost devices. However, limited resolution, power efficiency, and scalability hinder computational performance. While silicon offers high integration and efficiency, its rigidity and cost limit adaptability~\cite{mubarik2020printed}. In contrast, PE enables customizable circuits, but further advances are needed—especially for ML tasks constrained by power and area. These challenges demand new circuit architectures and arithmetic techniques suited to PE's unique properties~\cite{bleier2020printed}.

ML in PE enables real-time, on-device intelligence in resource-limited settings like wearable monitors and environmental sensors. Yet, ML classifiers in PE face challenges including computational cost, limited area, and high power consumption. Previous work explored bespoke classifiers that hardwire parameters to minimize overhead~\cite{chaidos2025bespoke}, improving efficiency but not addressing the major bottleneck—ADCs used for sensor data. Techniques such as approximate~\cite{armeniakos2023model,armeniakos2022cross} and stochastic computing~\cite{weller2021printed} aim to reduce this cost but may compromise accuracy, limiting their applicability. Thus, alternative solutions are needed to optimize printed ML classifiers further.

While algorithmic~\cite{armeniakos2022cross}, circuit~\cite{armeniakos2023model}, and logic-level~\cite{weller2021printed,armeniakos2023co} optimizations have been explored, no prior work investigates alternative arithmetic representations. This work extends current research by exploring arithmetic-level transformation using unary arithmetic, tailored to PE’s design flexibility. To our knowledge, this is the first study of unary implementations for printed ML models. Additionally, the impact of ADCs—major contributors to power and area use—has been largely overlooked~\cite{armeniakos2024sensor}. We show that unary arithmetic, with its computational simplicity and efficient routing, offers a promising direction. It enables direct sensor data handling, reducing ADC reliance and streamlining data conversion, thereby improving printed ML classifier efficiency.

To this end, we introduce a novel hybrid unary-binary architecture for efficient ML classifier implementations in PE. By altering arithmetic representation, we combine unary arithmetic's simplicity with binary's compactness, creating efficient, model-specific MLP hardware. Our approach achieves efficiency through: (i) reduced precision arithmetic, (ii) customized model training for hardware, and (iii) redesign of binary operations for unary implementation. This hybrid method supports low-power, area-efficient ML classifiers, addressing PE challenges and enabling advanced in-sensor computing with reduced resource demands.

\textbf{Our main contributions in this work can be summarized as follows:}

\begin{enumerate}[topsep=0pt,leftmargin=*]
    \item We propose a novel approach for efficiently performing multiplications, additions, and accumulations using unary arithmetic.
    \item We present a hybrid unary-binary architecture for MLPs and investigate its impact in designing digital ML classifiers.
    \item We present a holistic hardware-software co-design that includes layer-aware MLP training for multiplier-less, efficient implementation.
\end{enumerate}

\section{Proposed Co-design Methodology}
This section presents our co-design methodology for generating printed MLP classifiers that leverage our hybrid unary-binary architecture. First, we introduce the fully parallel unary core operations of MLPs, specifically multiplications and additions, and emphasize their efficiency benefits over traditional binary computation. Next, we present the hybrid unary-binary architecture of the MLP models and we finally describe our customized training method, which optimizes further area and power efficiency while preserving high accuracy.

\subsection{Fully-parallel Unary Operations}

In unary coding, the value of a unary stream is determined by the number of ones it contains. There are two primary ways to interpret this value:

\begin{enumerate}[topsep=0pt,leftmargin=*]
    \item \textbf{Integer-valued unary representation}: Each `1' in the stream counts as 1, so the total value is simply the count of `1's, regardless of the stream's length. For example:
    \begin{align}
    11000_\text{U} &= 010100_\text{U} = 00000011_\text{U} =2_{10}
\end{align}
    
    \item \textbf{Real-valued unary representation}: The value is expressed as the fraction of `1's over the total number of bits in the stream.  Here, each bit represents $1/n$ of the total value, where $n$ is the respective bit-width, and equivalent to the LSB in binary format. The representable range is $[0, 1]$ (unipolar), which can be extended to $[-1, 1]$ (bipolar with an additional sign bit) \cite{sim2017new}. For instance:
    \begin{align}
    0010_\text{U} &= 0001_\text{U} = 00010100_\text{U} = \frac{2}{8} = 0.25_{10}
    \end{align}    
\end{enumerate}

The arrangement of `1's in the stream defines two unary representations: (1) \textit{thermometer code}, where all `1's are grouped at one end (e.g., $1100_\text{U}=2_{10}$); and (2) \textit{rate encoding}, with randomly distributed `1's whose value depends solely on their frequency, commonly used in stochastic computing (e.g., $1010_\text{U}=2_{10}$).

\begin{figure}[t]
    \centering
    \includegraphics[width=1\linewidth]{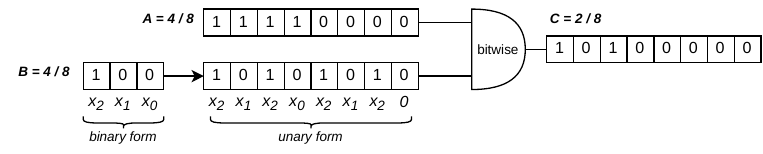}
    \caption{Two unary inputs (temporal encoding above, rate encoding below) undergo bitwise-AND for multiplication. Rate encoding requires offline binary-to-unary conversion (left).}
    \label{fig:unarymult}\vspace{-2ex}
\end{figure}

Transitioning from binary to unary computation requires redefining fundamental arithmetic operations. 
We focus on multiplication and addition, as these are the core operations for ML classifiers and not all operations benefit equally from unary representation~\cite{wu2020ugemm}.
Our unary constant multiplier extends an approximate stochastic multiplier concept through optimized hardware implementation. The design leverages bitwise-AND operations between equal-width input vectors, requiring complementary encodings: one input in temporal (thermometer) encoding and the other in rate encoding.

Fig.\ref{fig:unarymult} illustrates the proposed multiplication in bit-serial form. Although shown this way for clarity, the final design uses a fully parallel architecture, detailed later. Operands and outputs are represented as bitstreams, where the frequency of \textit{1s} defines the value. Input \textit{A} is treated as a unary number (its derivation is discussed later), while input \textit{B} follows unary unipolar encoding\cite{sim2017new}. This transformation is done offline, leveraging PE’s customization to hardwire values. The resulting constant unary multiplication produces a bitstream output with a maximum theoretical error of $1/N$, where $N$ is the bit-width~\cite{wu2020ugemm}. While inherently approximate, the result is sufficiently accurate for our purposes. Section~\ref{sec:res} analyzes this approximation’s impact on final classification accuracy.

Unary addition benefits from the non-positional nature of unary encoding, where all bits have equal weight. This allows a simple scheme where input bits are summed in parallel via binary addition (see subsection~\ref{sec:hybrid}). Multiple additions can be performed simultaneously by aggregating bits in one operation.

\begin{figure}[t]
    \centering
    \includegraphics[width=1\linewidth]{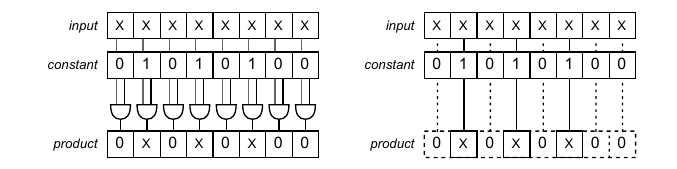}
    \caption{Bit-parallel unary multiplication. While each input bit theoretically requires AND operations with weight bits (left), zeros in the weight vector eliminate corresponding connections (dotted lines) and are omitted from the design. Active inputs route directly to outputs without AND gates (right).}
    \label{fig:parmul}\vspace{-2ex}
\end{figure}

The proposed architecture offers key hardware efficiency benefits. It eliminates the need for input encoders at ADCs, simplifies multiplication by replacing AND gates with efficient routing, and reduces operations on multi-bit operands. Offline analysis of hidden-layer weights using bitwise-OR identifies unused bits, enabling removal of corresponding comparators in the analog-to-unary converter. While individual gains are modest, together these optimizations meaningfully reduce area and power.
As highlighted, in hardwired MLPs with parallel unary inputs, all multiplication AND gates can be removed (see Fig.~\ref{fig:parmul})—ideal for the limited resources of printed circuits. A serial (temporal) unary format would not only require multi-cycle execution, unsuitable for PE, but also add significant hardware overhead, such as control logic and registers.

\subsection{Hybrid Unary-binary MLPs}\label{sec:hybrid}

\begin{figure}[t]
    \centering
    \includegraphics[width=0.7\linewidth]{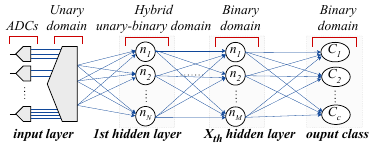}
    \caption{Overall architecture of our proposed hybrid unary-binary MLP}
    \label{fig:hybrid}\vspace{-2ex}
\end{figure}

In this section we introduce a hybrid MLP architecture that combines unary and binary computation. The first layer operates entirely in unary, while later layers use conventional fixed-point (fxp) binary logic, maintaining compatibility with existing hidden-to-output hardware, including the comparator tree (Fig.~\ref{fig:hybrid}).

\begin{figure}[t]
    \centering
    \includegraphics[width=0.95\linewidth]{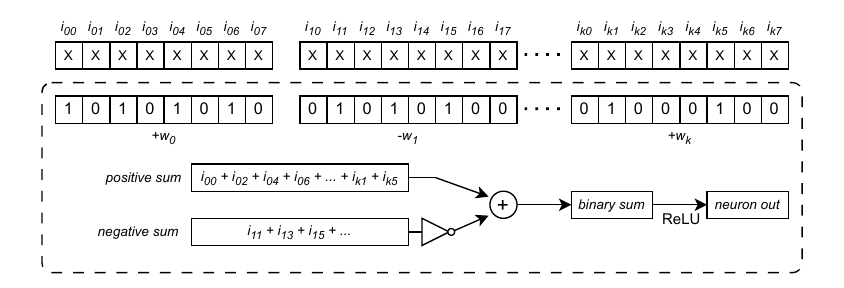}
    \caption{Illustrative example of a hybrid unary-binary neuron computation}
    \label{fig:neuron}\vspace{-2ex}
\end{figure}

Supplying the sensor input as a parallel unary word minimizes printed overheads. A flash ADC already yields a thermometer code, so removing its encoder converts the signal directly to the required unary format, reducing ADC area and power. Due to limited space, schematic details can be found in Fig.1 of~\cite{armeniakos2024sensor}.
Fig.~\ref{fig:neuron} illustrates a hybrid neuron: indices of hard-wired unary products are retrieved and summed in binary to produce the accumulation. In the second layer, positive and negative partial products are accumulated separately, then merged with a single signed add (plus bias), eliminating most multipliers without accuracy loss.
ReLU and the final comparator tree remain functionally unchanged, but the lower bit-width throughout the network further cuts area and power. The resulting design balances unary simplicity with binary compactness for highly efficient printed MLPs.

\subsection{Customized MLP training}

In this hybrid unary-binary design, hardware efficiency is primarily constrained by binary layer operations, so our training targets their optimization. Key to this is selecting MLP parameters that preserve high classification accuracy while enabling low-power operation—crucial for power-autonomous printed MLP classifiers.

We introduce a layer-aware training approach that maximizes area efficiency by quantizing weights to power-of-two values, allowing constant multiplications via simple wiring rather than logic, while retaining standard addition. The method balances accuracy loss and hardware gains to optimize both computation and performance.
Each binary layer is processed by adjusting neuron weights to their nearest power-of-two, followed by retraining for a fixed number of epochs $m$ (experimentally set to $m=2$) and re-evaluating accuracy. If the drop exceeds a threshold $T$, the change is reverted. This repeats until all weights are quantized or the accuracy constraint is breached. A valid solution always exists, as original weights can be retained in the worst case.
\section{Results \& analysis}\label{sec:res}

In this section, we evaluate our printed MLP classifiers by analyzing the hardware benefits of our hybrid unary-binary design and layer-aware MLP training. The evaluation is conducted on six datasets, as listed in Table~\ref{tab:baseline}. These datasets were chosen for two key reasons: (i) to enable direct comparisons with state-of-the-art methods~\cite{mubarik2020printed,armeniakos2023model,armeniakos2022cross}, and (ii) because they contain sensor-based inputs well-suited for printed applications. All datasets were sourced from the UCI ML repository. For training and testing, we normalize inputs to the [0,1] range and apply a random 70\%/30\% split. 
The digital circuits are analyzed using Synopsys Design Compiler and PrimeTime, with all circuits operating at 20Hz—matching the typical performance of target printed electronics (PE) applications~\cite{mubarik2020printed}.

\begin{table}[t]
\setlength\tabcolsep{4pt}
\renewcommand{\arraystretch}{1}
\caption{Evaluation of the baseline MLPs}
\begin{tabular}{cccccl}
\hline
\textbf{Dataset} & \textbf{Acc (\%)} & \textbf{\#Inputs} & \textbf{\#MAC ops} & \textbf{Area (cm$^2$)} & \textbf{Power (mW)} \\ \hline
\rowcolor[HTML]{EFEFEF} 
Cardio        & 88.6 & 21 & 72 & 12.8 & 40.8 \\
RedWine       & 57.1 & 11 & 34 & 8.2  & 26.9 \\
\rowcolor[HTML]{EFEFEF} 
WhiteWine     & 54.2 & 11 & 72 & 8.6  & 27.9 \\
Seeds         & 88.9 & 7  & 30 & 5.8  & 20.4 \\
\rowcolor[HTML]{EFEFEF} 
Vertebral 3C  & 77.4 & 6  & 27 & 4.9  & 17.6 \\
Balance Scale & 84.6 & 4  & 21 & 2.8  & 10.8 \\ \hline
\end{tabular}\label{tab:baseline}\vspace{-1ex}
\end{table}

\subsection{Comparison with exact binary baseline}


Table \ref{tab:baseline} summarizes the fully-parallel (one inference per cycle) baseline MLPs implemented exactly as in \cite{mubarik2020printed}. They use 3-bit inputs and 5-bit weights, achieving near-floating-point accuracy. Average area and power are \textcolor{black}{7.2 cm$^{2}$} and \textcolor{black}{24.1 mW}. These power demands exceed typical printed-battery limits (15–30 mW) \cite{mubarik2020printed}, so most designs cannot be self-powered.

\begin{figure}[t]
    \centering
    \includegraphics[scale=0.87]{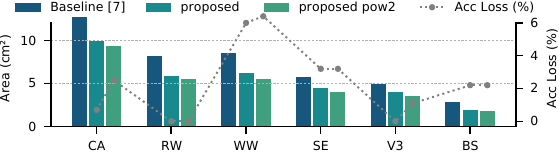}\\
    \includegraphics[scale=0.87]{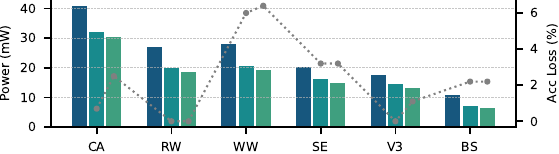}
    \caption{Area (a) and power (b) evaluation of our hybrid unary-binary design (proposed) and its enhanced version (training with pow2), compared to the baseline~\cite{mubarik2020printed}, along with the respective accuracy loss in \%}
    \label{fig:comp1}\vspace{-2ex}
\end{figure}

Fig.~\ref{fig:comp1} shows the area and power improvements achieved by our proposed hybrid unary-binary architecture and its enhanced version, which incorporates power-of-two training, compared to the baseline, i.e., the non-unary parallel architecture from~\cite{mubarik2020printed}.
As demonstrated, our hybrid unary-binary methodology results in significant hardware improvements, with average gains of \textcolor{black}{25\%} in area and \textcolor{black}{24\%} in power. Notably, these gains are achieved with an average accuracy loss of only \textcolor{black}{2\%}, and all datasets, except for WhiteWine, exhibit an accuracy loss of less than 3\%.
Additionally, we explore the impact of our customized power-of-two MLP training. On average, our approach successfully converts \textcolor{black}{33.6\%} of the weights into power-of-two values, resulting in an extra \textcolor{black}{8.5\%} and \textcolor{black}{7.9\%} gains in area and power, respectively, for less than \textcolor{black}{0.6\% additional} accuracy degradation.

In order to quantify the additional benefit obtained by eliminating unnecessary comparators in our bespoke ADCs, we define “input-bit utilization” as the share of actually used input bits; it averages 92\%. Further savings are possible by aligning zeros and ones across shared weights.

\subsection{Comparison with approximate state-of-the-art works}

\begin{figure}[t]
    \centering
    \includegraphics[scale=0.75]{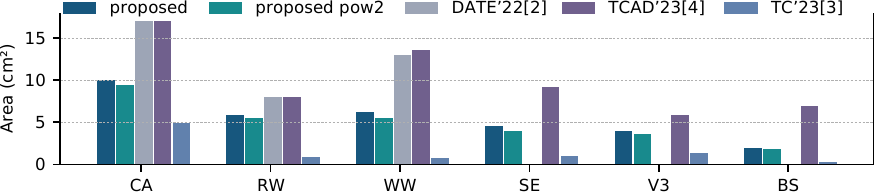}\\
    \includegraphics[scale=0.75]{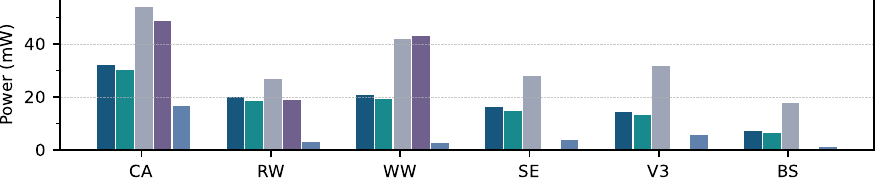}
    \caption{Area (a) and power (b) comparison of our proposed hybrid unary-binary designs vs state-of-the-art approximate binary MLPs of~\cite{armeniakos2022cross,armeniakos2023co,armeniakos2023model}.}
    \label{fig:sota}\vspace{-2ex}
\end{figure}

In Fig.\ref{fig:sota}, we compare our designs with state-of-the-art approximate printed MLPs\cite{armeniakos2022cross,armeniakos2023model,armeniakos2023co}. Since no prior work explores unary-encoded printed MLPs, we benchmark against existing techniques for binary implementations.

Our unary-binary MLPs outperform~\cite{armeniakos2022cross,armeniakos2023model} across all metrics—achieving on average \textcolor{black}{46\%} lower area and \textcolor{black}{39\%} lower power. While~\cite{armeniakos2023co} shows lower area and power, this is due to:
1) It omits ADC overheads. In binary, a single ADC can consume over $18\times$ the area of our worst-case pruned unary ADC~\cite{armeniakos2024sensor}, and this advantage scales with input count.
2) All weights in~\cite{armeniakos2023co} are power-of-two, which is unlikely for larger datasets. Our method is dataset-agnostic and more robust for complex models.
3) Wiring overhead is unaccounted for in prior work, whereas our design benefits from fewer input bits and logic, reducing routing demands.

Overall, our hybrid unary-binary method effectively targets printed applications of similar complexity to those in Table~\ref{tab:baseline}. For <5\% accuracy loss, it significantly reduces computational cost and power consumption—supporting classifiers under 30mW, suitable for self-powered operation~\cite{mubarik2020printed}.

\section{Conclusion}

This work presents a hybrid unary-binary approach for implementing MLP classifiers in printed electronics (PE), addressing large feature size constraints while leveraging PE’s flexibility and low-cost fabrication. By eliminating costly encoders and enabling multiplier-less computation, the design reduces hardware complexity. Architecture-aware training further improves area and power efficiency with minimal accuracy loss. Extensive evaluation confirms the method’s suitability for resource-constrained printed ML applications.

\begin{credits}
\subsubsection{\ackname} This project has received funding from the H2020 project EVOLVE, under grant agreement No 825061.

\end{credits}
%
%
%
%

\bibliographystyle{splncs04}
\bibliography{references}

\end{document}